\documentclass{article} 
\usepackage{tencent_ailab_tech_report}

\usepackage[T1]{fontenc} 
\usepackage[utf8]{inputenc} 

\usepackage{microtype}
\usepackage{hyperref}
\usepackage{url}
\usepackage{booktabs}
\usepackage[table]{xcolor}
\usepackage{colortbl}
\definecolor{darkblue}{rgb}{0, 0, 0.5}
\hypersetup{colorlinks=true, citecolor=darkblue, linkcolor=darkblue, urlcolor=darkblue}

\usepackage{graphicx}
\usepackage{enumitem}
\usepackage{pgfplots}
\usepackage{tabularx}
\usepackage{caption}
\usepackage{subcaption}
\usepackage{bbm}
\usepackage{url}
\usepackage{amsmath} 
\usepackage{pifont}
\usepackage{arydshln}
\usepackage{makecell}
\usepackage{multirow} 
\usepackage{algorithm}
\usepackage{wrapfig}
\usepackage{amssymb}

\usepackage{algpseudocode}
\usepackage{lscape}       
\usepackage{adjustbox}    

\usepackage[most]{tcolorbox}

\definecolor{lightgray}{gray}{0.93}
\definecolor{lightblue}{RGB}{220,235,247}

\newtcolorbox{bluebox}[1][]{
  enhanced,
  colframe=blue!40!gray,
  colback=white,
  coltitle=white,
  colbacktitle=blue!40!gray,
  width=\linewidth,
  arc=2mm,
  auto outer arc,
  boxrule=0.5pt,
  left=10pt,
  right=10pt,
  drop shadow={black!50!white},
  top=10pt,
  bottom=10pt,
  title={#1}, 
  fonttitle=\bfseries,
  title code={\node[rounded corners, fill=blue!75!black, draw=none, text=white] at (frame.title) {\textbf{#1}};}, 
  attach boxed title to top center={yshift=-2mm},
  boxed title style={sharp corners, size=small}
}

\definecolor{blockA}{RGB}{248,249,250} 
\definecolor{blockB}{RGB}{241,243,245} 
\definecolor{oursRow}{RGB}{237,247,237} 

\newcolumntype{A}{>{\columncolor{blockA}}c}
\newcolumntype{B}{>{\columncolor{blockB}}c}

\usepackage[most]{tcolorbox}

\newtcolorbox{keyfindingbox}[1]{%
  enhanced,
  colback=gray!5,          
  colframe=gray!60!black,  
  colbacktitle=gray!40!black, 
  coltitle=white,          
  fonttitle=\bfseries,     
  boxed title style={sharp corners},
  sharp corners,
  title=#1
}

\definecolor{lightgray}{gray}{0.93}
\definecolor{lightblue}{RGB}{220,235,247}

\DeclareUnicodeCharacter{211D}{\ensuremath{\mathbb{R}}}
\DeclareUnicodeCharacter{2124}{\ensuremath{\mathbb{Z}}}
\DeclareUnicodeCharacter{21A6}{\ensuremath{\mapsto}}

\DeclareUnicodeCharacter{2264}{\ensuremath{\leq}}
\DeclareUnicodeCharacter{2265}{\ensuremath{\geq}}
\DeclareUnicodeCharacter{2192}{\ensuremath{\rightarrow}}
\DeclareUnicodeCharacter{2194}{\ensuremath{\leftrightarrow}}
\DeclareUnicodeCharacter{2227}{\ensuremath{\wedge}}
\DeclareUnicodeCharacter{22A2}{\ensuremath{\vdash}}
\DeclareUnicodeCharacter{2080}{\ensuremath{_0}}
\DeclareUnicodeCharacter{2081}{\ensuremath{_1}}
\DeclareUnicodeCharacter{2082}{\ensuremath{_2}}
\DeclareUnicodeCharacter{2083}{\ensuremath{_3}}
\DeclareUnicodeCharacter{2084}{\ensuremath{_4}}
\DeclareUnicodeCharacter{2085}{\ensuremath{_5}}
\DeclareUnicodeCharacter{2086}{\ensuremath{_6}}
\DeclareUnicodeCharacter{2087}{\ensuremath{_7}}
\DeclareUnicodeCharacter{2088}{\ensuremath{_8}}
\DeclareUnicodeCharacter{2089}{\ensuremath{_9}}
\DeclareUnicodeCharacter{2200}{\ensuremath{\forall}}
\DeclareUnicodeCharacter{2203}{\ensuremath{\exists}}
\DeclareUnicodeCharacter{2228}{\ensuremath{\lor}}
\DeclareUnicodeCharacter{2260}{\ensuremath{\neq}}
\DeclareUnicodeCharacter{230B}{\ensuremath{\rfloor}} 
\DeclareUnicodeCharacter{2223}{\ensuremath{\mid}}    
\DeclareUnicodeCharacter{230A}{\ensuremath{\lfloor}} 
\DeclareUnicodeCharacter{2090}{\ensuremath{_a}}  
\DeclareUnicodeCharacter{2091}{\ensuremath{_e}}  
\DeclareUnicodeCharacter{2092}{\ensuremath{_o}}  
\DeclareUnicodeCharacter{2093}{\ensuremath{_x}}  
\DeclareUnicodeCharacter{2211}{\ensuremath{\sum}}        
\DeclareUnicodeCharacter{2115}{\ensuremath{\mathbb{N}}}  

\DeclareUnicodeCharacter{2208}{\ensuremath{\in}} 

\usepackage{etoolbox}

\title{%
\hspace*{-0.5cm}
  \begin{minipage}{1.0\textwidth} 
Can LLMs Guide Their Own Exploration? Gradient-Guided Reinforcement Learning for LLM Reasoning
  \end{minipage}
}

\author{
Zhenwen Liang$^{1}$,
Sidi Lu$^{1}$, Wenhao Yu$^1$, Kishan Panaganti$^1$,
Yujun Zhou$^{1,2}$,
Haitao Mi$^1$,\\
\hspace{0.2cm}\textbf{Dong Yu$^1$} \vspace{1em}\\
$^1$Tencent AI Lab, 
$^2$University of Notre Dame \\
Correspondence to: \texttt{zhenwzliang@global.tencent.com}}

\colmfinalcopy
\begin{document}

\maketitle

\begin{abstract}
Reinforcement learning has become essential for strengthening the reasoning abilities of large language models, yet current exploration mechanisms remain fundamentally misaligned with how these models actually learn. Entropy bonuses and external semantic comparators encourage surface-level variation but offer no guarantee that sampled trajectories differ in the \emph{update directions} that shape optimization. We propose \emph{G\textsuperscript{2}RL}, a gradient-guided reinforcement learning framework in which exploration is driven not by external heuristics but by the model’s own first-order update geometry. For each response, G\textsuperscript{2}RL constructs a sequence-level feature from the model’s final-layer sensitivity—obtainable at negligible cost from a standard forward pass—and measures how each trajectory would reshape the policy by comparing these features within a sampled group. Trajectories that introduce novel gradient directions receive a bounded multiplicative reward scaler, while redundant or off-manifold updates are deemphasized, yielding a self-referential exploration signal that is naturally aligned with PPO-style stability and KL control. Across math and general reasoning benchmarks (\textsc{MATH500}, \textsc{AMC}, \textsc{AIME24}, \textsc{AIME25}, \textsc{GPQA}, \textsc{MMLUpro}) on Qwen3-base 1.7B/4B models, G\textsuperscript{2}RL consistently improves pass@1, maj@16, and pass@k over entropy-based GRPO and external-embedding methods. Analyzing the induced geometry, we find that G\textsuperscript{2}RL expands exploration into substantially more orthogonal—and often opposing—gradient directions while maintaining semantic coherence, revealing that a policy’s own update space provides a far more faithful and effective basis for guiding exploration in LLM RL.
\end{abstract}


\section{Introduction}

Reinforcement learning (RL) has become a central mechanism for improving the reasoning and decision-making abilities of large language models (LLMs), extending their capabilities beyond supervised finetuning \citep{christiano2017deep, ouyang2022training, rafailov2023direct}. Yet, despite this progress, exploration in LLM RL remains fundamentally underdeveloped. Current exploration strategies—entropy bonuses, outcome rarity, or external semantic comparators—are all driven by \emph{signals extrinsic to the model}. They encourage the policy to sample more widely, but they do so without regard for the model’s own update structure. As a result, exploration often becomes diffuse, misaligned, or fragile under sparse reward signals, especially when supervision is binary or verifiable \citep{sutton1998reinforcement, auer2002finite, kakade2002approximately}.  

Entropy increases randomness but is oblivious to whether two responses induce meaningfully different directions. External similarity models based on semantic embeddings provide sequence-level contrast but operate in representation spaces that differ from the model's internal geometry. A trajectory may appear “novel’’ semantically yet offer no new gradient information for the policy; conversely, trajectories essential for improving the model's reasoning may be penalized for superficial semantic similarity. In short, existing methods guide exploration through lenses \emph{external} to the policy, producing an enduring mismatch between exploration signals and the optimization dynamics that actually govern learning.

\begin{figure}[t]
    \centering
    \includegraphics[width=0.85\linewidth]{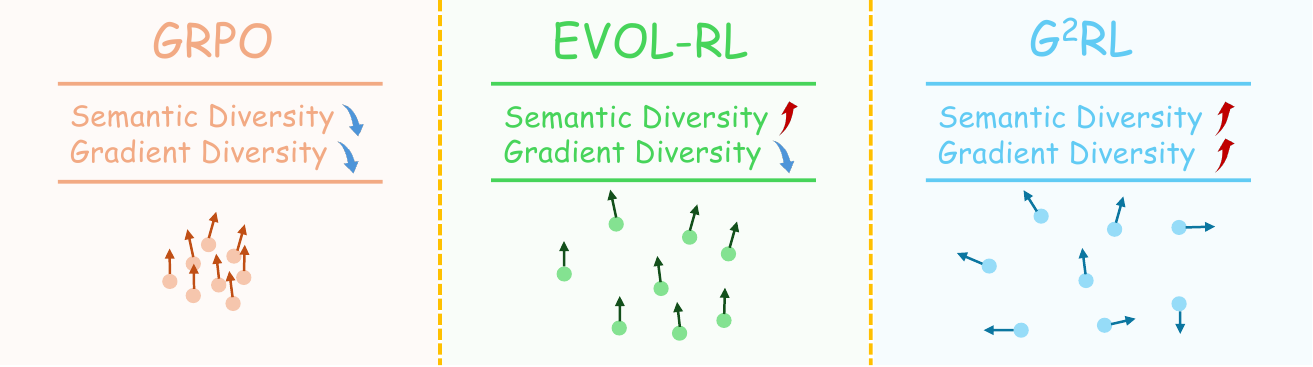}
    \caption{
    Comparison of the characteristics of GRPO, EVOL-RL, and G\textsuperscript{2}RL. 
    The dispersion of points indicates semantic variety, while arrows represent gradient directions. 
    Only G\textsuperscript{2}RL explicitly encourages exploration aligned with the model’s intrinsic update geometry.
    }
    \vspace{-5mm}
    \label{fig:method_comparison}
    
\end{figure}

This motivates a more principled question:  
\emph{Can an LLM learn to explore by examining how each trajectory would reshape its own parameters, rather than relying on external proxies?}  
Such an approach would shift exploration from being externally imposed to being \emph{self-guided}: trajectories are preferred when they meaningfully expand the policy’s own update directions, and discouraged when they contribute redundant or uninformative gradients.

We propose \emph{G\textsuperscript{2}RL}, a Gradient-guided reinforcement learning framework that realizes this idea. Instead of encouraging diversity in the output space, G\textsuperscript{2}RL encourages exploration directly in the policy’s \emph{update geometry}. For each generated response, we construct a sequence-level feature derived from the model’s own first-order sensitivity at the final layer—available at negligible cost from a standard forward pass. This feature summarizes how the trajectory would steer the model’s output distribution through its gradients. By comparing these features within a group of candidates, we obtain a \emph{policy-intrinsic exploration score}: responses that introduce new gradient directions are upweighted through a bounded reward scaler, whereas responses that merely reinforce existing update directions receive less emphasis.

This gradient-guided mechanism offers three conceptual advantages.  
First, it grounds exploration in the same geometry that governs optimization, eliminating the semantic–optimization mismatch inherent in external comparators.  
Second, it provides a self-referential criterion: the model explores what \emph{it} stands to learn from, not what appears diverse to an auxiliary encoder.  
Third, the construction integrates seamlessly into GRPO \citep{shao2024deepseekmath}, requiring no extra backward passes.

We evaluate G\textsuperscript{2}RL across math and general reasoning benchmarks and multiple LLM scales, finding consistent improvements in \texttt{pass@1}, \texttt{maj@16}, and \texttt{pass@k}. Beyond raw accuracy, G\textsuperscript{2}RL produces richer reasoning trajectories and more meaningful gradient dispersion, supporting the central hypothesis that exploration should be guided not by external heuristics but by the policy’s own update geometry.

\textbf{Contributions.}
\begin{itemize}[leftmargin=10pt]
    \item We introduce G\textsuperscript{2}RL, a gradient-guided RL method that defines exploration through the model’s own first-order update geometry, avoiding reliance on entropy or external semantic encoders.
    \item We propose a bounded, groupwise reward-scaling mechanism that emphasizes trajectories offering novel gradient directions while preserving optimization stability.
    \item Experiments on math and general reasoning tasks demonstrate that G\textsuperscript{2}RL systematically improves single-sample accuracy and multi-sample coverage, achieving healthier and more structurally meaningful exploration dynamics.
\end{itemize}

\section{Method}

We introduce G\textsuperscript{2}RL, a gradient-guided reinforcement learning method for large language models (LLMs) that augments group-relative policy optimization (GRPO) with an exploration signal computed in the policy’s own update space. Rather than rewarding diversity in output space, G\textsuperscript{2}RL adjusts the contribution of each trajectory according to how it reshapes the model’s gradient directions. This section formalizes the setting, recalls GRPO, derives the gradient features, defines a groupwise gradient-guided exploration score, and integrates it into a stable PPO-style objective with KL control. All symbols are summarized as they appear.

\subsection{Preliminaries and Group-Relative Policy Optimization}

Let $\mathcal{X}$ be a set of prompts and $\mathcal{Y}$ the response space. For $x\in\mathcal{X}$, a response is the token sequence $y=(y_1,\dots,y_L)\in\mathcal{Y}$. An autoregressive LLM with parameters $\theta$ defines
\begin{equation}
p_\theta(y\mid x) \;=\; \prod_{t=1}^L p_\theta\!\left(y_t \mid x, y_{<t}\right).
\label{eq:llm}
\end{equation}
We generate, for each prompt $x$, a \emph{group} of $m$ candidate responses $\{y^{(i)}\}_{i=1}^m$ from a fixed behavior policy $\pi_{\theta^{\mathrm{old}}}$ (autoregressive or nucleus sampling). Each response receives a \emph{base} scalar reward $r^{(i)}\in\mathbb{R}$, e.g., a verifiable pass/fail or a task-specific score.

GRPO \citep{shao2024deepseekmath} dispenses with a learned critic and estimates groupwise advantages by standardizing rewards within the group:
\begin{equation}
\bar r \;=\; \frac{1}{m}\sum_{i=1}^m r^{(i)}, \qquad
s_r \;=\; \sqrt{\frac{1}{m}\sum_{i=1}^m \bigl(r^{(i)}-\bar r\bigr)^2 + \varepsilon},
\qquad
A^{(i)} \;=\; \frac{r^{(i)} - \bar r}{s_r},
\label{eq:grpo-adv}
\end{equation}
with $\varepsilon>0$ for numerical stability. Let
\begin{equation}
\rho^{(i)} \;=\; \frac{\pi_\theta\!\left(y^{(i)}\!\mid x\right)}{\pi_{\theta^{\mathrm{old}}}\!\left(y^{(i)}\!\mid x\right)},
\qquad
\mathrm{clip}\!\left(u;1-\epsilon,1+\epsilon\right)\;=\;\min\!\bigl\{\max\{u,1-\epsilon\},\,1+\epsilon\bigr\},
\label{eq:ratio-clip}
\end{equation}
and let $D_{\mathrm{KL}}\!\left(\pi_\theta(\cdot\mid x)\,\Vert\,\pi_{\mathrm{ref}}(\cdot\mid x)\right)$ be a per-prompt KL penalty to a reference policy (e.g., the SFT model). The GRPO objective is
\begin{equation}
\mathcal{J}_{\mathrm{GRPO}}(\theta)
\;=\;
\mathbb{E}_{x,\{y^{(i)}\}\sim \pi_{\theta^{\mathrm{old}}}}
\Biggl[
\frac{1}{m}\sum_{i=1}^m
\min\!\Bigl(\rho^{(i)} A^{(i)},\,
\mathrm{clip}\!\bigl(\rho^{(i)};1-\epsilon,1+\epsilon\bigr) A^{(i)}\Bigr)
\;-\; \beta\, D_{\mathrm{KL}}\!\left(\pi_\theta\!\Vert\!\pi_{\mathrm{ref}}\right)
\Biggr].
\label{eq:grpo-obj}
\end{equation}
The clipping term stabilizes policy updates, while the KL regularizer prevents drift from the reference policy. G\textsuperscript{2}RL modifies only the way advantages are constructed, leaving PPO-style clipping and KL control unchanged.

\subsection{Gradient Features: Policy-Referential Sensitivity}

Let $h_t\in\mathbb{R}^{d_h}$ denote the final-layer hidden state at time $t$, and let the LM head be linear:
\begin{equation}
z_t \;=\; W^\top h_t + b \;\in\; \mathbb{R}^{V},
\qquad
p_t \;=\; \mathrm{softmax}\!\left(\frac{z_t}{T}\right),
\label{eq:head-softmax}
\end{equation}
where $W\in\mathbb{R}^{V\times d_h}$, $b\in\mathbb{R}^{V}$, vocabulary size $V$, and temperature $T>0$. Writing $e(y_t)\in\{0,1\}^V$ for the one-hot of the realized token, the standard score-function identity yields the exact token-level gradient with respect to $h_t$:
\begin{equation}
\frac{\partial \log p_\theta\!\left(y_t\mid x,y_{<t}\right)}{\partial h_t}
\;=\;
\frac{1}{T}\, W\Bigl(e(y_t) - p_t\Bigr)
\;\in\; \mathbb{R}^{d_h}.
\label{eq:token-grad}
\end{equation}
Define the token residual $r_t := e(y_t)-p_t$ and the token feature
\begin{equation}
\phi_t \;:=\; W r_t \in \mathbb{R}^{d_h}.
\label{eq:token-feature}
\end{equation}
Up to the constant $1/T$, $\phi_t$ is the exact first-order sensitivity of the log-likelihood to perturbations at $h_t$. Aggregating over the response produces a sequence-level feature
\begin{equation}
\Phi(x,y) \;=\; \sum_{t=1}^{L} \tilde{\alpha}_t\, \phi_t,
\qquad
\tilde{\alpha}_t \;=\; \frac{\alpha_t}{\sum_{s=1}^L \alpha_s + \varepsilon},
\label{eq:seq-feature}
\end{equation}
with default uniform weights $\alpha_t\equiv 1$ and masking for non-response tokens if applicable.

\emph{Interpretation.}
By \eqref{eq:head-softmax}--\eqref{eq:token-grad}, any infinitesimal perturbation
$\delta h_t$ induces
$\delta \log p_\theta(y_t\mid\cdot) \approx \langle \phi_t/T,\, \delta h_t\rangle$,  
so the sequence feature $\Phi(x,y)$ summarizes, to first order, how the response
$(x,y)$ would steer the model’s output distribution through its final-layer
representation.
As detailed in Appendix~\ref{app:geometry}, the full parameter gradient along a trajectory factors through this feature: for every layer $k$ there exists a trajectory-dependent linear operator $\mathcal{L}_k(x,y)$ such that
\[
\nabla_{\theta_k} \ell(x,y) = \frac{1}{T}\,\mathcal{L}_k(x,y)\,\Phi(x,y).
\]
Thus, all upstream updates lie in linear images of the same sequence feature
$\Phi(x,y)$, and angular relations between responses in $\Phi$-space are propagated—up to layerwise linear transforms—to the actual optimization directions. We use these features not because the last layer dominates all upstream effects, but because it forms the unique first-order sensitivity bottleneck through which trajectory-specific information must pass, and it is available from the forward pass without any extra backpropagation. This makes cosine geometry on $\Phi$ a principled and computationally cheap proxy for comparing how different trajectories guide the policy’s updates.

\subsection{Groupwise Gradient-Guided Exploration Score}

Given a group $\{\Phi^{(i)}\}_{i=1}^m$, we first unit-normalize the
sequence features and define pairwise cosine similarities:
\begin{equation}
\hat{\Phi}^{(i)} \;=\;
\frac{\Phi^{(i)}}{\|\Phi^{(i)}\|_2 + \varepsilon},
\qquad
S_{ij} \;=\; \bigl\langle \hat{\Phi}^{(i)}, \hat{\Phi}^{(j)} \bigr\rangle
\;\in\;[-1,1].
\label{eq:cosine}
\end{equation}

Next, we construct \emph{reward-weighted} coefficients
\begin{equation}
w_{ij}
\;=\;
\frac{\exp\!\bigl(r^{(j)}\bigr)\,\mathbb{1}\{j\neq i\}}
     {\sum_{k\neq i}\exp\!\bigl(r^{(k)}\bigr)+\varepsilon},
\label{eq:weights}
\end{equation}
which form a probability distribution over the other candidates in the
group: $w_{ij}\ge 0$ and $\sum_{j\neq i} w_{ij}=1$.
Higher-reward responses therefore act as more important reference
directions when we assess the contribution of $y^{(i)}$.

Using these ingredients, we define a bounded, scale-invariant
\emph{gradient-guided exploration score}:
\begin{equation}
\nu^{(i)}
\;=\;
\sqrt{
  \max\!\Bigl(
    1 - \sum_{j\neq i} w_{ij}\, S_{ij}^2,\; 0
  \Bigr)
}
\;\in\;[0,1].
\label{eq:novelty}
\end{equation}

Intuitively, the term
$\sum_{j\neq i} w_{ij}\, S_{ij}^2$ measures how well the direction
$\hat{\Phi}^{(i)}$ can be “explained’’ by a weighted combination of
the other responses' gradient directions. If $\hat{\Phi}^{(i)}$ is
almost parallel to several high-reward peers, the weighted
squared cosine similarities $S_{ij}^2$ are large and their sum
approaches~1; in that case, $1 - \sum_{j\neq i} w_{ij} S_{ij}^2$ is small and
$\nu^{(i)}$ is close to~0, indicating that $y^{(i)}$ contributes little new information to the update geometry.
Conversely, if $\hat{\Phi}^{(i)}$ is nearly orthogonal to most
high-reward responses, all $S_{ij}^2$ are small, the weighted sum
is far below~1, and $\nu^{(i)}$ is close to~1.
Thus $\nu^{(i)}$ can be read as the “remaining’’ component of the
update direction for $y^{(i)}$ that is not captured by the dominant,
high-reward directions in the group.
Because we work with unit vectors in \eqref{eq:cosine}, this score
is invariant to any common rescaling of the underlying features
$\Phi^{(i)}$.

\subsection{Gradient-Guided Reward Shaping}

We convert the exploration score into a multiplicative reward factor that preserves optimization stability. Let $\bar{\nu}^{(i)}$ denote a bounded, monotone transformation of $\nu^{(i)}$ (e.g., min–max normalization within the group). We define
\begin{equation}
f\!\left(\nu^{(i)}, r^{(i)}\right)
\;=\;
1 + \lambda\, \left(r^{(i)}\right)\, \bar{\nu}^{(i)},
\qquad
\lambda \in [0,\lambda_{\max}],
\label{eq:scale}
\end{equation}
and the gradient-guided reward is
\begin{equation}
\tilde{r}^{(i)} \;=\; r^{(i)} \cdot f\!\left(\nu^{(i)}, r^{(i)}\right).
\label{eq:shaped-reward}
\end{equation}

This formulation induces an asymmetric effect. For correct responses ($r^{(i)}=1$), high exploration score boosts the reward, prioritizing trajectories that follow successful yet geometrically distinct update directions over redundant repetitions. For incorrect responses ($r^{(i)}=-1$), high exploration score \emph{amplifies the penalty} ($\tilde{r}^{(i)} < -1$), while low exploration score (high similarity to correct peers) \emph{mitigates the penalty} ($\tilde{r}^{(i)} > -1$). An incorrect response with low exploration score has its gradient feature $\Phi^{(i)}$ aligned with the subspace of correct solutions, suggesting a “near-miss’’ whose update direction is still informative; by penalizing these less, the policy is encouraged to stay within a plausible reasoning manifold. Conversely, an incorrect response with high exploration score is nearly orthogonal to successful trajectories, indicating off-manifold or hallucinated behavior that should be suppressed.

\subsection{Practical Reward Rescaling}

In all our experiments the base rewards are binary,
$r^{(i)}\in\{-1,1\}$, indicating incorrect vs.\ correct responses.
The gradient-guided factor in \eqref{eq:scale} modifies these rewards but we keep the
overall scale tightly controlled. Concretely, we instantiate the mapping
$f(\nu^{(i)}, r^{(i)})$ so that it is bounded and monotone in the normalized
exploration score, and we clip the shaped reward in \eqref{eq:shaped-reward} to a
fixed interval:
\[
\tilde{r}^{(i)}
\;\leftarrow\;
\mathrm{clip}\bigl(\tilde{r}^{(i)};\,-c,\,c\bigr),
\qquad c = 3.
\]
Since $r^{(i)}\in\{-1,1\}$, this guarantees that the effective reward magnitude
seen by the policy always satisfies $|\tilde{r}^{(i)}|\le 3$, so the gradient-guided term can at most moderately up- or down-weight individual samples. This keeps the advantage scale stable while still allowing the exploration signal to reshape the relative weighting of candidates within each group.

\subsection{Computation and Implementation}

\textit{Token features without extra backprop.}
Equation~\eqref{eq:token-grad} depends only on quantities from the forward pass. Computing $\phi_t = W r_t$ can be implemented as
\begin{equation}
\phi_t \;=\; W\,e(y_t)\;-\;W\,p_t \;=\; W_{(:,y_t)} \;-\; \mathbb{E}_{v\sim p_t}\!\bigl[\,W_{(:,v)}\,\bigr],
\label{eq:efficient}
\end{equation}
i.e., a column gather and a matrix–vector product with $p_t$. Aggregating $\phi_t$ over $t$ yields $\Phi(x,y)$ with negligible overhead relative to the softmax and log-prob computations already performed. In practice, G\textsuperscript{2}RL is therefore a drop-in modification of GRPO that replaces groupwise advantages by gradient-guided ones while leaving the rest of the RL pipeline unchanged.

\section{Experiments}
\label{sec:experiments}

We evaluate the proposed G\textsuperscript{2}RL on two Qwen3 base models (1.7B and 4B). Our gradient-guided exploration term is applied as reward shaping within the groupwise standardization.
We use the \textsc{MATH} training set yielding 7.5k training problems \citep{hendrycks2021measuring}. A rule-based verifier provides both the RL reward signal during training and the correctness oracle at evaluation time.

\subsection{Benchmarks and Metrics}

We report results on \textsc{AIME24}, \textsc{AIME25}, \textsc{MATH500}, \textsc{AMC}, and \textsc{GPQA}. For each prompt we report pass@1, maj@16 and pass@16. 
\(\text{pass@k}\) is the fraction of prompts for which at least one of the \(k\) samples is verified correct. 
\(\text{maj@16}\) is the majority-vote accuracy over 16 samples: we select the most frequent final answer string among the 16 and verify that single prediction (ties broken uniformly). All numbers are percentages; bold indicates the best within each block.

\subsection{Main Results}

\begin{table}[h]
\centering
\small
\caption{Main results on MATH500, AMC, AIME24, AIME25 with Qwen3-1.7B-Base and Qwen3-4B-Base backbone.}
\label{tab:main-all-transpose-reorder}
\resizebox{\textwidth}{!}{
\begin{tabular}{l BBB AAA BBB AAA}
\toprule
 & \multicolumn{3}{c}{\cellcolor{blockB}MATH500} & \multicolumn{3}{c}{\cellcolor{blockA}AMC} &
   \multicolumn{3}{c}{\cellcolor{blockB}AIME24} & \multicolumn{3}{c}{\cellcolor{blockA}AIME25} \\
\cmidrule(lr){2-4}\cmidrule(lr){5-7}\cmidrule(lr){8-10}\cmidrule(lr){11-13}
Model & \cellcolor{blockB}pass@1 & \cellcolor{blockB}maj@16 & \cellcolor{blockB}pass@16
      & \cellcolor{blockA}pass@1 & \cellcolor{blockA}maj@16 & \cellcolor{blockA}pass@16
      & \cellcolor{blockB}pass@1 & \cellcolor{blockB}maj@16 & \cellcolor{blockB}pass@16
      & \cellcolor{blockA}pass@1 & \cellcolor{blockA}maj@16 & \cellcolor{blockA}pass@16 \\
\midrule
\multicolumn{13}{l}{\textbf{Qwen3-1.7B-Base}} \\
GRPO
& 63.5 & 73.2 & 86.6  & 31.2 & 42.0 & 68.1  & 7.5 & 14.8 & 24.4  & 4.6 & 6.8 & 22.2 \\
Entropy Bonus
& 64.3 & 74.1 & 86.7  & 32.2 & 42.3 & 65.7  & 9.6 & 17.2 & 23.3  & 4.6 & 6.9 & 24.5 \\
EVOL-RL
& 64.3 & 73.7 & 86.9  & 32.2 & 43.9 & 66.2  & 8.4 & 15.8 & 27.3  & 5.3 & 7.9 & 21.6 \\
\rowcolor{oursRow}
\textbf{G\textsuperscript{2}RL}
& \textbf{66.2} & \textbf{76.8} & \textbf{88.7}
& \textbf{33.9} & \textbf{44.8} & \textbf{68.5}
& \textbf{10.1} & \textbf{17.4} & \textbf{28.0}
& \textbf{7.5} & \textbf{11.4} & \textbf{24.8} \\
\midrule
\multicolumn{13}{l}{\textbf{Qwen3-4B-Base}} \\
GRPO
& 76.9 & 81.6 & 90.8  & 47.9 & 56.2 & 75.1  & 12.4 & 18.2 & 31.1  & 10.0 & 16.2 & 32.5 \\
Entropy Bonus
& 79.0 & 87.2 & 93.2  & 50.5 & 63.7 & 79.5  & 17.8 & 25.4 & 40.0  & 16.1 & 24.4 & 41.5 \\
EVOL-RL
& 80.0 & 87.7 & 93.5  & 50.9 & 62.0 & 81.9  & 19.4 & 28.2 & 42.4  & 17.5 & 23.9 & 39.8 \\
\rowcolor{oursRow}
\textbf{G\textsuperscript{2}RL (Ours)}
& \textbf{80.8} & \textbf{87.8} & \textbf{93.6}
& \textbf{52.3} & \textbf{63.8} & \textbf{82.0}
& \textbf{19.9} & \textbf{28.7} & \textbf{43.8}
& \textbf{20.1} & \textbf{29.0} & \textbf{45.0} \\
\bottomrule
\end{tabular}
}
\end{table}

On 1.7B, the gradient-guided variant G\textsuperscript{2}RL improves both single-try quality and multi-sample coverage across all datasets, indicating that exploration in the model’s own update space leads to more effective use of the sampling budget. The gains are largest on \textsc{AIME25}, where \(\text{pass@1}\) rises to 7.5 and \(\text{maj@16}\) to 11.4, indicating that G\textsuperscript{2}RL encourages useful optimization-space exploration rather than mere noise. On \textsc{MATH500}, improvements are consistent through \(\text{pass@16}\) (88.7 vs.\ 86.9 for the strongest baseline). The only case where a baseline slightly edges out G\textsuperscript{2}RL is \(\text{AMC}\,\text{pass@16}\) (68.1 vs.\ 68.5), while G\textsuperscript{2}RL leads on the remaining metrics for that dataset.

The 4B results amplify these trends. On \textsc{AIME25}, \(\text{pass@1}\) improves from 17.5 (best baseline) to 20.1 and \(\text{maj@16}\) from 23.9 to 29.0. On \textsc{MATH500}, gains are smaller but consistent across all metrics, reaching \(\text{pass@16}=93.6\). On \textsc{AMC}, G\textsuperscript{2}RL maintains the best \(\text{pass@1}\), \(\text{maj@16}\), and \(\text{pass@16}\), indicating stronger sample efficiency and coverage at scale.

\begin{table}[t]
\centering
\small
\caption{General reasoning results on the backbone of Qwen3-4B-Base. Left: GPQA (multiple-choice) with full sampling metrics. Right: MMLUpro micro-average pass@1. All numbers are percentages; best in bold.}
\label{tab:gen-reasoning}
\resizebox{\textwidth}{!}{
\begin{tabular}{@{}p{0.66\textwidth} p{0.32\textwidth}@{}}
\multicolumn{1}{c}{\textbf{GPQA (4B)}} & \multicolumn{1}{c}{\textbf{MMLUpro (4B, pass@1)}} \\
\\[-0.7em]
\begin{tabular}{l r r r r}
\hline
Method & pass@1 & maj@16 & pass@16 & pass@32 \\
\hline
GRPO         & 37.2 & 39.2 & 81.2 & 85.2 \\
Entropy Bonus    & 37.8 & 42.8 & 88.6 & 92.6 \\
EVOL-RL          & 37.4 & 42.1 & 88.9 & \textbf{93.7} \\
\textbf{G\textsuperscript{2}RL (Ours)} & \textbf{38.7} & \textbf{44.0} & \textbf{89.2} & 93.5 \\
\hline
\end{tabular}
&
\begin{tabular}{l r}
\hline
Method & Micro Avg. \\
\hline
GRPO          & 56.15 \\
Entropy Bonus               & 57.14 \\
EVOL-RL                     & 57.17 \\
\textbf{G\textsuperscript{2}RL (Ours)} & \textbf{58.47} \\
\hline
\end{tabular}
\end{tabular}}
\end{table}

Across both model sizes, G\textsuperscript{2}RL improves \(\text{pass@1}\) on every dataset, showing that gradient-guided exploration shifts probability mass toward higher-quality solutions. Gains in \(\text{pass@16}\) demonstrate better coverage of distinct correct modes, especially on challenging splits like \textsc{AIME25}.

\subsection{Analysis on General Reasoning Tasks}

We assess generalization on two broad-coverage reasoning benchmarks using the 4B model. On \textsc{GPQA}, which is four-option multiple-choice, sampling benefits are pronounced: \(\text{pass@k}\) rises with \(k\). Our method improves single-try quality and consensus (\(\text{pass@1}=38.7\), \(\text{maj@16}=44.0\)) and achieves the best \(\text{pass@16}=89.2\); at \(\text{pass@32}\) the task is near-saturated and EVOL-RL is marginally higher (93.7 vs.\ 93.5). On the larger and more diverse MMLUpro, we report \(\text{pass@1}\) only, our approach attains a higher micro-average (58.47) than EVOL-RL (57.17), entropy bonus (57.14), and GRPO (56.15). These results indicate that the gradient-guided exploration signal improves generalization beyond math-style settings, raising both single-try accuracy and useful coverage without auxiliary encoders.

\subsection{Training dynamics}

\begin{center}
\begin{keyfindingbox}{Key Finding 1}
G\textsuperscript{2}RL achieves the fastest and highest gains in accuracy
and response length while keeping entropy moderate; entropy bonuses mainly
inflate entropy and token count, and EVOL-RL shows healthy but ultimately
weaker improvements due to its externally defined exploration signal.
\end{keyfindingbox}
\end{center}

\begin{figure}[h]
  \centering
  \includegraphics[width=\linewidth]{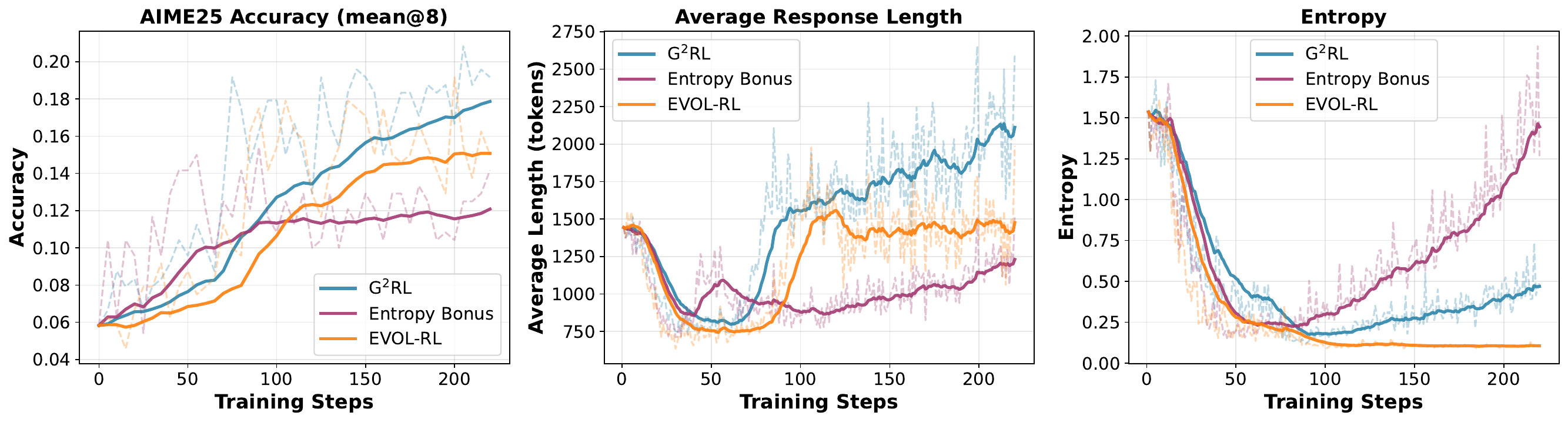}
  \caption{Training dynamics on \textsc{AIME25}: $\mathrm{mean@8}$ accuracy,
  average response length, and entropy for GRPO, Entropy Bonus, EVOL-RL, and G\textsuperscript{2}RL.}
  \label{fig:training-dynamics}
\end{figure}

To understand how different exploration mechanisms influence learning, we compare
training dynamics for three methods:
\textbf{Entropy Bonus}, \textbf{EVOL-RL}, and \textbf{G\textsuperscript{2}RL}.
We track \textsc{AIME25} $\mathrm{mean@8}$ accuracy, average response length, and
output entropy over training steps (Figure~\ref{fig:training-dynamics}).

\paragraph{Accuracy and response length.}
G\textsuperscript{2}RL shows the steepest and most stable improvement in $\mathrm{mean@8}$:
its accuracy curve rises quickly and plateaus at the highest level among all
methods. EVOL-RL improves more smoothly than entropy bonus,
but still converges below G\textsuperscript{2}RL. In parallel, G\textsuperscript{2}RL drives a rapid early
increase in average response length, indicating that the model quickly learns
to produce richer, more structured reasoning rather than merely recycling short
patterns.
For Entropy Bonus, length is more volatile and less predictive of accuracy.

\paragraph{Entropy as a noisy proxy for exploration.}
Entropy Bonus unsurprisingly produces the largest increase in output entropy,
but this growth is only weakly coupled to accuracy: entropy continues to rise
even when accuracy saturates, suggesting that many additional tokens are
uninformative for solving the task. G\textsuperscript{2}RL, in contrast, yields a moderate but
\emph{aligned} entropy increase: entropy rises together with both accuracy and
response length, reflecting exploration that contributes to useful reasoning.
EVOL-RL behaves in between these extremes.

\paragraph{Why G\textsuperscript{2}RL outperforms EVOL-RL.}
Both EVOL-RL and G\textsuperscript{2}RL maintain “healthy’’ curves where accuracy and length
co-evolve without obvious instability. The key difference lies in how exploration signal
is defined. EVOL-RL relies on an external encoder whose similarity geometry is
only loosely tied to the current policy, and thus cannot perfectly adapt to its
evolving representation. G\textsuperscript{2}RL instead measures exploration pressure directly in the policy’s gradient feature space, keeping the exploration signal aligned with the actual update directions and yielding faster learning and higher final accuracy.

\subsection{Analysis of Exploration Geometry}
\label{sec:diversity-analysis}

\begin{center}
\begin{keyfindingbox}{Key Finding 2}
G\textsuperscript{2}RL alters the optimization landscape, increasing the ratio of opposing gradient directions (negative similarity) by nearly \textbf{5$\times$} compared to GRPO. Crucially, we observe a distinct misalignment between semantic-space and gradient-space geometry: external embeddings fail to capture the structural orthogonality that drives efficient exploration.
\end{keyfindingbox}
\end{center}

\begin{figure}[h]
  \centering
  \includegraphics[width=\linewidth]{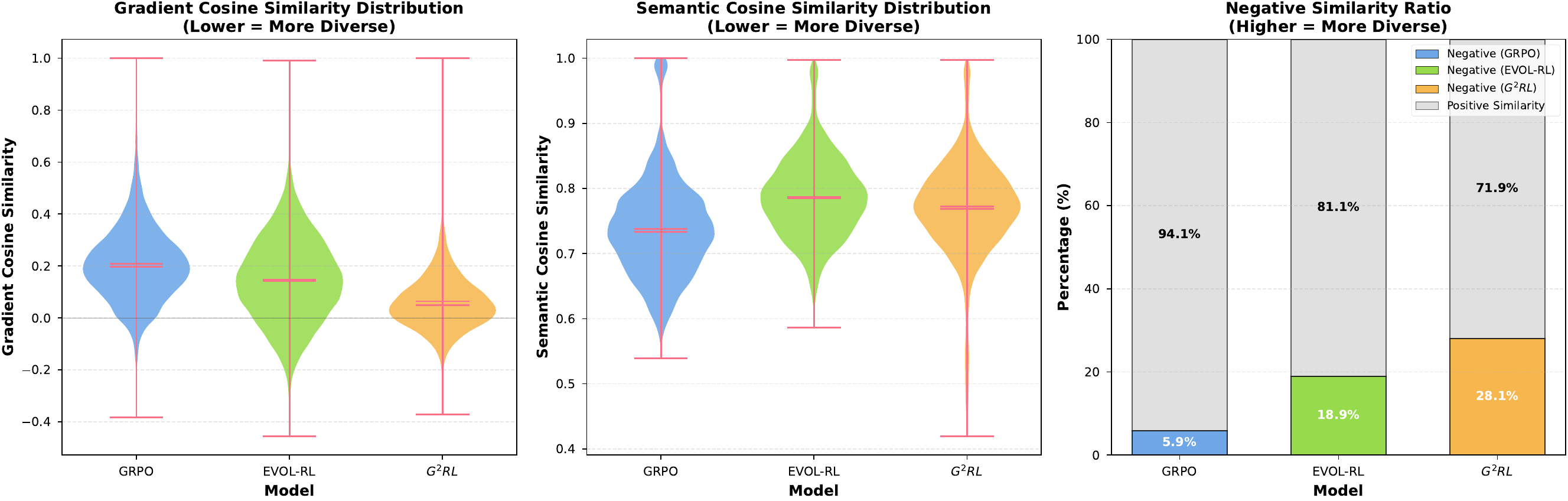}
  \caption{Exploration-geometry analysis on \textsc{AIME25}. \textbf{Left:} Distribution of pairwise gradient-space cosine similarity. G\textsuperscript{2}RL significantly shifts the distribution toward zero. \textbf{Middle:} Distribution of semantic-space cosine similarity. G\textsuperscript{2}RL maintains high semantic coherence despite much more diverse gradient geometry. \textbf{Right:} Ratio of negative similarity pairs. G\textsuperscript{2}RL generates nearly \textbf{5$\times$} more orthogonal or opposing optimization directions than vanilla GRPO.}
  \label{fig:diversity-analysis}
\end{figure}

To investigate the underlying mechanism of our method, we conducted a controlled analysis of the exploration geometry induced by different training strategies in both the policy's native gradient space and an external semantic space.
We sampled 8 responses for each of 30 randomly selected problems from the AIME25 validation set using three models: vanilla GRPO, EVOL-RL, and G\textsuperscript{2}RL. We measured pairwise similarity using two metrics: \textit{Gradient Geometry} (based on the policy's own update features $\Phi$) and \textit{Semantic Similarity} (based on an external embedding model).

\paragraph{Gradient Geometry and Orthogonality.}
As shown in Figure~\ref{fig:diversity-analysis} (Left), G\textsuperscript{2}RL shifts the distribution of pairwise gradient similarities significantly toward zero. While vanilla GRPO responses exhibit high collinearity (mean cosine similarity $0.208$), G\textsuperscript{2}RL reduces this to $0.064$, indicating a much broader coverage of the optimization space.
Most notably, we analyze the \textit{Negative Similarity Ratio} (Figure~\ref{fig:diversity-analysis}, Right), which tracks response pairs pointing in opposing directions. Vanilla GRPO produces only $5.91\%$ negative pairs. In contrast, G\textsuperscript{2}RL boosts this to $28.09\%$. This confirms that our gradient-guided exploration mechanism successfully drives the policy toward structurally distinct reasoning paths that offer complementary gradient information, rather than merely rephrasing similar solutions.

\paragraph{Misalignment in Semantic Space.}
The results in semantic space (Figure~\ref{fig:diversity-analysis}, Middle) reveal a critical insight. Vanilla GRPO actually yields lower semantic similarity ($0.738$) than G\textsuperscript{2}RL ($0.769$).
A naive interpretation might suggest GRPO is "more diverse." However, given GRPO's inferior performance, this likely reflects incoherent or off-manifold exploration. G\textsuperscript{2}RL maintains higher semantic consistency (staying on-topic and coherent) while simultaneously maximizing gradient orthogonality. This discrepancy shows that external semantic embeddings are an unreliable proxy for RL exploration: they may penalize subtle but update-relevant reasoning variations that are valuable in the policy’s own gradient space.

\section{Discussion: The Geometry of Efficient Exploration}

\paragraph{Breaking Collinearity via Orthogonal Gradients.}
Standard GRPO treats all correct responses uniformly, assigning them identical positive advantages. Our analysis of gradient geometry reveals a structural flaw in this approach: under vanilla GRPO, successful trajectories exhibit high gradient collinearity (mean similarity 0.21). This implies that nominally ``diverse'' samples are often redundant in optimization space, pushing parameters along the same dominant direction and accelerating mode collapse. G\textsuperscript{2}RL fundamentally alters this dynamic. By explicitly guiding exploration in the policy's own sensitivity space, it does not merely spread samples out in output space; it encourages them to be \emph{functionally orthogonal} in update space. Our experiments show a 5$\times$ increase in negative similarity pairs compared to the baseline, suggesting that G\textsuperscript{2}RL actively selects trajectories that provide \emph{complementary} gradient information—effectively counterbalancing over-represented directions of the dominant mode and keeping the optimization landscape flat enough to permit continued, stable exploration.

\paragraph{Semantic vs.\ Optimization Geometry.}
A critical insight from our comparison with EVOL-RL is the misalignment between human-intelligible semantic variation and optimizer-relevant structural variation. Intuitively, one might expect lower semantic similarity to correlate with better exploration. However, our results show the opposite: G\textsuperscript{2}RL maintains \emph{higher} semantic consistency (0.77) than vanilla GRPO (0.74) while achieving drastically \emph{lower} gradient similarity. This indicates that external embedding models are deceptive proxies for RL: they may reward surface-level changes (phrasing, irrelevant tangents) that contribute little to learning, or penalize subtle but high-value reasoning shifts that matter for optimization. G\textsuperscript{2}RL succeeds because it operates directly in the \emph{policy’s intrinsic gradient geometry}: it encourages variations that maximize the geometric difference in the parameter update step, regardless of whether those variations appear ``semantically distinct'' to an external encoder. In doing so, it effectively decouples \emph{useful exploration} from \emph{random noise} in output space.

\paragraph{Credit Assignment within Correctness Classes.}
Finally, G\textsuperscript{2}RL addresses a subtle credit assignment ambiguity in sparse-reward settings. In binary-reward math tasks, the optimizer cannot distinguish between a fragile, pattern-matched solution and a robust, principled one if both happen to be correct. By modulating the reward with gradient-guided weights, G\textsuperscript{2}RL implements a \emph{dynamic re-weighting} scheme: it down-weights the ``easy,'' repetitive paths whose gradient directions are already crowded, and amplifies the signal from rarer trajectories that open new update directions. This allows the model to continue extracting learning signal from correct answers long after a standard policy gradient would have saturated on a single mode, and it does so by leveraging the model’s own update geometry as the reference for what constitutes informative exploration.

\section{Related Work}

\textbf{Exploration and diversity in reinforcement learning.}
A long line of work encourages exploration by maximizing entropy, which improves robustness and prevents premature convergence by discouraging early overcommitment to narrow modes \citep{haarnoja2018soft}. Quality–diversity (QD) methods such as MAP-Elites extend this idea by simultaneously maintaining performance and behavioral diversity, yielding repertoires of high-quality yet distinct solutions \citep{mouret2015illuminating, pugh2016quality}. In unsupervised RL, DIAYN maximizes mutual information between latent skills and states to acquire a set of diverse policies without external rewards \citep{eysenbach2018diversity}. Taken together, these approaches demonstrate that structured exploration is not merely auxiliary but central to stable learning and broad generalization. However, they typically reason about exploration in terms of entropy or behavior space, rather than in terms of the geometry of parameter updates.

\textbf{Exploration-aware RL formulations for LLMs.}
Recent work on LLM reinforcement learning makes exploration explicit in the training objective. \citet{song2025outcome} diagnose diversity collapse under majority-style training and propose outcome-based exploration, which assigns bonuses to rare outcomes (historically or within-batch) to recover coverage without sacrificing accuracy. DARLING jointly optimizes a learned diversity signal with task reward, showing gains on both non-verifiable and competition-math settings \citep{li2025jointly}. Label-free formulations also emphasize exploration to avert collapse: EVOL-RL couples majority-based selection for stability with novelty-aware rewards for exploration, embodying a variation–selection principle \citep{zhou2025evolving}, while RESTRAIN turns spurious majority signals into penalties for overconfident, low-consistency rollouts, enabling self-driven RL that maintains healthy variability without gold labels \citep{yu2025restrain}. 

Our work, G\textsuperscript{2}RL, is aligned with this trajectory in that it modifies the RL objective to reshape exploration, but it differs in how the exploration signal is defined. Instead of relying on entropy or external embedding spaces to measure diversity, G\textsuperscript{2}RL computes policy-referential similarity in the model’s own gradient feature space, avoiding auxiliary encoders and aligning the exploration signal directly with the policy’s update geometry.

\section{Conclusion}
G\textsuperscript{2}RL offers a principled and practical framework for guiding exploration in large language models by anchoring it to the policy’s own update geometry rather than to entropy or external semantic encoders. By deriving a sequence-level representation from last-layer gradient sensitivity and integrating it as a bounded, reward-coupled weighting within GRPO, the method selectively amplifies correct trajectories that introduce new optimization directions and suppresses incoherent ones, breaking gradient collinearity without compromising training stability. Across math and general reasoning benchmarks and two Qwen3-base model scales, this simple modification yields consistent gains in pass@1, maj@16, and pass@k, produces healthier training dynamics, and increases the prevalence of genuinely orthogonal gradient pairs by five-fold—while maintaining coherent, on-topic outputs. These results support a broader perspective: efficient exploration in LLM reinforcement learning is achieved not by inflating entropy or superficial semantic dispersion, but by shaping the geometry of the optimization landscape through gradient-guided, policy-intrinsic signals.

\bibliography{colm2024_conference}
\bibliographystyle{colm2024_conference}

\newpage
\appendix
\section{Appendix}

\subsection{Theoretical Motivation: Diversity in the Optimization Landscape}
\label{app:geometry}

\subsubsection{First-Order Sensitivity at the Last Layer}

For a generated trajectory $(x,y)$, the contribution of token $t$ to the
log-likelihood gradient in last-layer space is
\begin{equation}
    \nabla_{h_t} \ell_t
    = \frac{1}{T} W \bigl(e(y_t) - p_t\bigr)
    =: \frac{1}{T}\,\phi_t,
    \label{eq:lastlayer-grad}
\end{equation}
where $p_t = \mathrm{softmax}(W^\top h_t/T)$.
Collecting all token-level gradients yields the sequence-level sensitivity
\begin{equation}
    G(h_{1:L})
    := 
    \bigl(\nabla_{h_1}\ell,\dots,\nabla_{h_L}\ell\bigr)
    =
    \frac{1}{T}(\phi_1,\dots,\phi_L).
    \label{eq:sequence-sensitivity}
\end{equation}

\subsubsection{Layerwise Factorization of Backpropagation}
\label{sec:linear_bottleneck}

The network computation graph is
\begin{equation}
    h_{k+1} = f_{k+1}(h_k;\theta_k), 
    \qquad k = 0,\dots,L-1.
\end{equation}
During backpropagation, the parameters are fixed.  
Define the Jacobians
\begin{equation}
    J_{k+1}^{(h)} := 
    \frac{\partial h_{k+1}}{\partial h_k},
    \qquad
    J_{k+1}^{(\theta)}
    := 
    \frac{\partial h_{k+1}}{\partial \theta_k}.
    \label{eq:jacobians}
\end{equation}
Applying the chain rule repeatedly gives the per-token upstream gradient
\begin{equation}
    \nabla_{\theta_k} \ell_t
    =
    \bigl(J_{k+1}^{(\theta)}\bigr)^\top
    \Bigl(J_{k+2}^{(h)}\Bigr)^\top
    \cdots
    \Bigl(J_{L}^{(h)}\Bigr)^\top
    \nabla_{h_L} \ell_t.
    \label{eq:layerwise-chainrule}
\end{equation}

Equation \eqref{eq:layerwise-chainrule} can be organized by defining the cumulative
transition operator
\begin{equation}
    \mathcal{T}_{k\to L}
    :=
    J_{k+2}^{(h)}
    J_{k+3}^{(h)}
    \cdots
    J_{L}^{(h)}
    \label{eq:transition-operator}
\end{equation}
(with $\mathcal{T}_{L\to L} = I$), so that
\begin{equation}
    \nabla_{\theta_k} \ell_t
    =
    \bigl(J_{k+1}^{(\theta)}\bigr)^\top
    \mathcal{T}_{k\to L}^\top
    \nabla_{h_L} \ell_t.
    \label{eq:compact-chain}
\end{equation}

Substituting \eqref{eq:lastlayer-grad} yields the explicit upstream factorization:
\begin{equation}
    \nabla_{\theta_k} \ell_t
    =
    \frac{1}{T}
    \underbrace{
      \bigl(J_{k+1}^{(\theta)}\bigr)^\top
      \mathcal{T}_{k\to L}^\top
    }_{\mathcal{L}_k(x,y)}
    \phi_t.
    \label{eq:upstream-linear-image}
\end{equation}
Here 
\begin{equation}
    \mathcal{L}_k(x,y)
    :=
    \left(J_{k+1}^{(\theta)}\right)^\top
    \mathcal{T}_{k\to L}^\top
\end{equation}
is a trajectory-dependent linear operator determined entirely by the forward pass
activations.

Equation \eqref{eq:upstream-linear-image} exhibits a complete structural decomposition:
\begin{equation}
    \boxed{
        \nabla_{\theta_k} \ell_t
        \;\in\;
        \operatorname{Im}\bigl(\mathcal{L}_k(x,y)\bigr)
        \cdot \phi_t.
    }
    \label{eq:image-subspace}
\end{equation}

\paragraph{Two structural sources of diversity.}
From \eqref{eq:image-subspace}, diversity in parameter updates across responses can
arise only from:
\begin{align}
    &\textbf{(i) Variation in }\phi_t 
    &&\text{(differences in last-layer sensitivity to tokens)},\\
    &\textbf{(ii) Variation in }\mathcal{L}_k(x,y)
    &&\text{(differences in intermediate activations and Jacobians)}.
\end{align}

\paragraph{Why shaping $\Phi$ is principled.}
For a response $y$, define a sequence-level feature such as
\begin{equation}
    \Phi(x,y)
    = \sum_{t=1}^L \alpha_t \phi_t,
\end{equation}
so that the full upstream gradient satisfies
\begin{equation}
    \nabla_{\theta_k} \ell(x,y)
    =
    \frac{1}{T}\,
    \mathcal{L}_k(x,y)\,
    \Phi(x,y).
    \label{eq:upstream-factorization-final}
\end{equation}
Thus all optimization signals factor through the same structural pipeline:
\[
\Phi(x,y)
\;\longrightarrow\;
\mathcal{L}_k(x,y)
\;\longrightarrow\;
\nabla_{\theta_k} \ell(x,y).
\]

Shaping the geometry of $\Phi$ therefore shapes the geometry of the entire
family of upstream gradients in \eqref{eq:upstream-factorization-final}.
In particular, promoting angular dispersion among $\Phi(x,y)$ encourages the resulting updates to explore a broader subspace of the parameter space, without
making assumptions on the Jacobians---because $\Phi$ is the unique quantity
through which all upstream gradients must factor linearly.

\subsection{Training Configuration.} 

We conduct our experiments on two recent open-source base models: \textbf{Qwen3-1.7B-Base} and \textbf{Qwen3-4B-Base}. Our training process is implemented using the GRPO algorithm. To ensure that the model has sufficient capacity for complex, multi-step reasoning, we set the maximum response length to 8k and 12k tokens during generation for 1.7B and 4B models, respectively. To guide the model's reasoning process, we utilize the system prompt from SimpleRL-Zoo \citep{zeng2025simplerl}:

\begin{bluebox}[System Prompt]
\texttt{
Please reason step by step, and put your final answer within \textbackslash boxed\{\}.
}
\end{bluebox}

Our training hyper-parameters are:

\begin{table}[h]
\centering
\caption{General hyper-parameters for RL training.}
\label{tab:general_hyperparams}
\begin{tabular}{lc}
\toprule
\textbf{Hyperparameter} & \textbf{Value} \\
\midrule
Train Batch Size & 16 \\
PPO Mini-Batch Size & 1 (effective size of 32) \\
PPO Micro-Batch Size & 2 \\
Rollouts & 16 \\
Generation Temperature & 1.0 \\
Validation Temperature & 0.8 \\
Learning Rate & 5e-7 \\
Use KL Loss & True \\
KL Loss Coefficient & 0.001 \\
\bottomrule
\end{tabular}
\end{table}

\end{document}